\documentclass[10pt,twocolumn,letterpaper]{article}

\usepackage{cvpr}              

\usepackage{graphicx}
\usepackage{amsmath}
\usepackage{amssymb}
\usepackage{booktabs}
\usepackage{ulem}
\usepackage{url}
\graphicspath{ {pic/} }

\usepackage[pagebackref,breaklinks,colorlinks]{hyperref}

\usepackage[capitalize]{cleveref}
\crefname{section}{Sec.}{Secs.}
\Crefname{section}{Section}{Sections}
\Crefname{table}{Table}{Tables}
\crefname{table}{Tab.}{Tabs.}

\def\cvprPaperID{*****} 
\def\confName{CVPR}
\def\confYear{2023}

\begin{document}

\title{A Creative Industry Image Generation Dataset Based on Captions and Sketches}

\author{Yuejia Xiang\footnotemark[1] \footnotemark[2]\\
{\tt\small xiangyuejia@bytedance.com}
\and
Chuanhao Lv\footnotemark[1]\\
{\tt\small 2001782@stu.neu.edu.cn}
\and
Qingdazhu Liu\\
{\tt\small 787673451@qq.com}
\and
Xiaocui Yang\\
{\tt\small yangxiaocui@stumail.neu.edu.cn}
\and
Bo Liu\\
{\tt\small 736343658@qq.com}
\and
Meizhi Ju\footnotemark[2]\\
{\tt\small meizhi.ju@outlook.com}
}

\maketitle

\renewcommand{\thefootnote}{\fnsymbol{footnote}}
\footnotetext[1]{These authors contributed equally to this work.} 
\footnotetext[2]{Corresponding authors.} 

\begin{abstract}
Most image generation methods are difficult to precisely control the properties of the generated images, such as structure, scale, shape, etc., which limits its large-scale application in creative industries such as conceptual design and graphic design, and so on. Using the prompt and the sketch is a practical solution for controllability. Existing datasets lack either prompt or sketch and are not designed for the creative industry.
Here is the main contribution of our work. a) This is the first dataset that covers the 4 most important areas of creative industry domains and is labeled with prompt and sketch. b) We provide multiple reference images in the test set and fine-grained scores for each reference which are useful for measurement. c) We apply two state-of-the-art models to our dataset and then find some shortcomings, such as the prompt is more highly valued than the sketch.
\end{abstract}

\section{Introduction}
Recently, image generation methods have drawn much attention from the community \cite{goodfellow2020generative, isola2017image, wang2018high}. But the controllability of these methods still improvements and the generated pictures often deviate from the expectation of art creators in the creative industry domain \cite{vojtovivc2015creative, banks2009after} (cf. Section \ref{sec:industry_domain}). Therefore, art creators need to make heavy modifications of the generated image to meet the required object, shape, color, structure, and atmosphere. The controllability limits the widespread application of image generation methods in the creative industry domain \cite{anantrasirichai2021artificial, rethinking2022media}.
To address the controllability of image generation, many methods have been proposed by previous work, such as prompt, sketch, segment, style picture, etc. \cite{liu2020sketch, chen2020deepfacedrawing, huang2021multimodal, schaldenbrand2022styleclipdraw}. After careful consideration, we chose only prompt and sketch to control image generation with the following reasons: a) prompt and sketch are the most understandable and most accessible material for art creators, b) prompt and sketch can represent the art creator's intention relatively completely\cite{jenkins1993importance, ekwaro2016uncertainty}.

Existing datasets \cite{abouelnaga2016cifar, deng2009imagenet, sharma2018conceptual, changpinyo2021conceptual, kakaobrain2022coyo-700m, schuhmann2021laion, li2019photo, sangkloy2016sketchy, karras2017progressive, zhang2021m6} lack either prompt or sketch and they are not designed for the creative industry domain, where the data is classified into 4 classes: concept design, graphic design, 3D-CG, and outdoor design. To solve these problems, we build an image generation dataset: a) Selecting images that belong to the creative industry domain from text-visual datasets such as CC12M \cite{changpinyo2021conceptual}. b) Generating sketches for the images by CLIPasso\cite{vinker2022clipasso}. c) Providing multiple sketches for each image and manually removing sketches of substandard quality. Note that if N sketches are kept for one image, then N cases are constructed, each of which contains one sketch. As shown in Table \ref{all_dataset_domain}, our dataset is more suitable for the creative industry domain compared to previous work. 

\begin{figure*}[t!]
\centering
\includegraphics[scale=0.20]{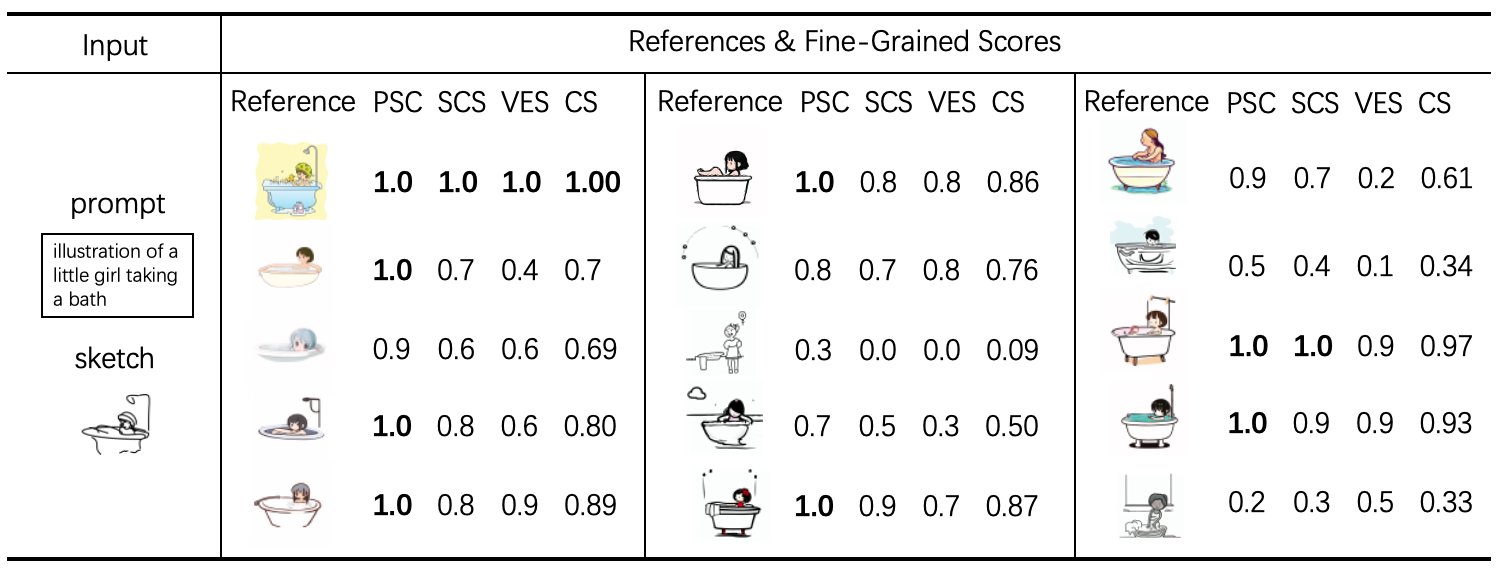}
\centering
\caption{An example of the test data. Each case in the test data generally has 5 to 20 reference images, which consist of one original image used to generate the sketch (named the golden reference image, which is the reference image located at the top left corner) and several images generated by SD \cite{rombach2022high} and NovelAI \cite{NovelAI}. All fine-grained scores (PCS, SCS, VES) for the golden reference image were set to 1.0, and the other reference images were obtained by manual annotation. Finally, A composite score (CS) is calculated by weighted average.
\label{test_data_example}}
\end{figure*}

\label{sec:intro}
\begin{table}
\centering
\caption{Comparing our dataset with previous work. PDID means the Percentage of Data in the creative Industry Domain. NICC means the Number of creative Industry Category Coverage. Although tags are a type of prompt, to distinguish tags from long text descriptions, we mark the dataset with only tags as No Prompt in this table.}
\label{all_dataset_domain}
\small
\begin{tabular}{lrrrr}\toprule[2pt]
Dataset & Prompt & Sketch & PDID & NICC \\
\midrule[1pt]
CIFAR-10 \cite{abouelnaga2016cifar} & N & N & 2\%  & 4\\
ImageNet \cite{deng2009imagenet} & N & N & 1\% & 4\\
Danbooru2021 \cite{danbooru2021} & N & N & 100\% & 1\\
CC12M \cite{changpinyo2021conceptual} & Y & N & 1\% & 4\\
COYO-700M \cite{kakaobrain2022coyo-700m} & Y & N & 1\% & 4\\
LAION-400M \cite{schuhmann2021laion} & Y & N & 1\% & 4\\
Photo-Sketching \cite{li2019photo} & N & Y & 1\% & 1\\
Sketchy \cite{sangkloy2016sketchy} & N & Y & 3\% & 4\\
MM-CelebA-HQ \cite{karras2017progressive} & Y & Y & 0\% & 0\\
M2C-Fashion \cite{zhang2021m6} & Y & Y & 100\% & 1\\
\midrule[1pt]
Our & Y & Y & 100\% & 4\\
\bottomrule[1pt]
\end{tabular}  
\end{table}

\begin{table}
\centering
\caption{Count the number of metadata of each type in the train set and the test set of our dataset. }
\label{our_dataset_statistic}
\begin{tabular}{lrr}\toprule[2pt]
Metadata & Train Set & Test Set\\
\midrule[1pt]
Image & 23155 & 500\\
De-duplicated Images & 13227 & 293\\
Prompt & 23155 & 500\\
Sketch & 23155 & 500\\
Reference & 23155 & 6928\\
Fine-Grained Score & 0 & 27712\\
\bottomrule[1pt]
\end{tabular}  
\end{table}

To more accurately measure the effect of the image generation model on our dataset, as shown in Fig \ref{test_data_example}, we enhance the test set: a) Inspired by Blue \cite{papineni2002bleu}, we provide multiple reference images for each test case (consisting of 1 prompt and 1 sketch). b) To more accurately assess the quality of images, we manually label each reference image with fine-grained scores, such as PCS (prompt consistency score), SCS (sketch consistency score), and VES (visual expressions score), and then obtain a CS (comprehensive score) by a weighted average method (cf. Section \ref{sec:evaluation_metrics}). c) To demonstrate the value of multiple reference images and fine-grained scores, we propose a new evaluation method named wFID (weighted Fréchet Inception Distance, cf. Section \ref{sec:weighted_fid}) whose evaluation results are closer to the manual evaluation results than FID \cite{heusel2017gans}.

Finally, we obtained a training set with 1 prompt, 1 sketch, and 1 golden reference image for each case and a test set with 1 prompt, 1 sketch, 1 golden reference image, and multiple reference images generated by the model for each case in the test set. The statistical information of our dataset is shown in Table \ref{our_dataset_statistic}.

\begin{figure*}[t!]
\centering
\includegraphics[scale=0.25]{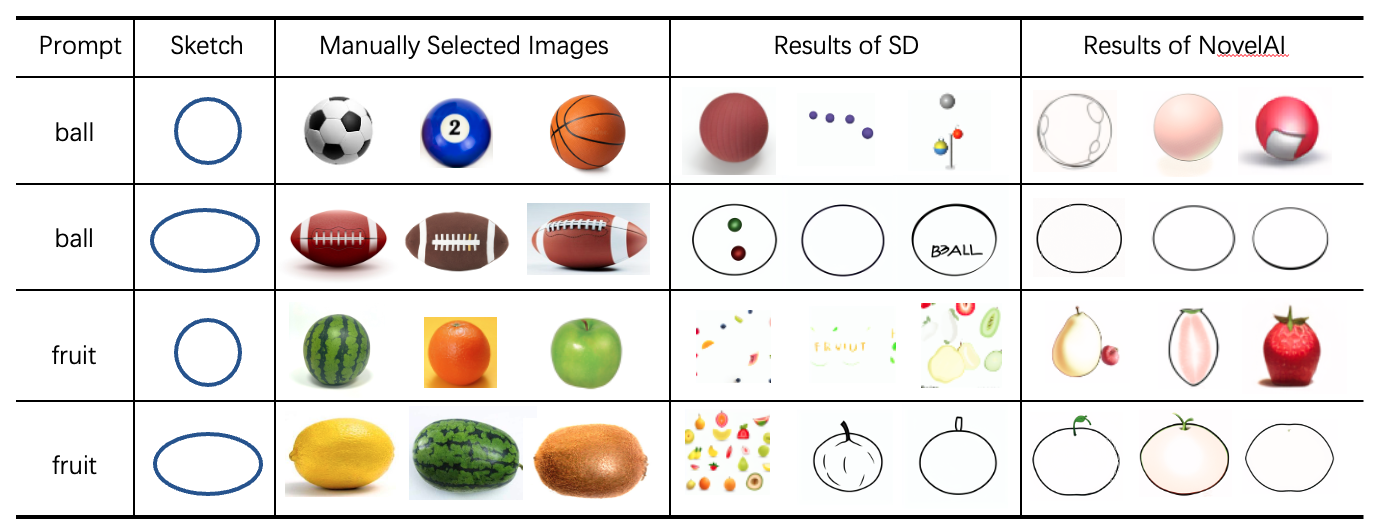}
\centering
\caption{Both prompt and sketch can be used to control image generation. In this figure, we control the material by prompt and the shape by sketch. For each pair composed of one prompt and one sketch we show 9 images, of which 3 are manually selected, 3 are generated by the Stable Diffusion (SD) \cite{rombach2022high} model, and 3 are generated by the NovelAI \cite{NovelAI} model. It can be found that these two models follow the prompt constraint better but follow the sketch constraint worse.
\label{prompt_and_sketch}}
\end{figure*}

We then experiment with our data on two state-of-the-art image generation models that support both prompt and sketch as the input and we observe 8 valuable phenomena (cf. Section \ref{sec:Discussion}), for example, as shown in Fig. \ref{prompt_and_sketch}, we find that both prompt and sketch can control the images generated by models, but the model generated images have a big gap in controllability compared to the images selected by a human.

\paragraph{Our Contribution}
\begin{itemize}
    \item This is the first dataset that covers the 4 most important areas of creative industry domains and is labeled with prompt and sketch. We choose prompt and sketch because prompt and sketch are the most understandable and accessible material for art creators. And we ensure that every sketch in the dataset is high quality by manual annotation. We believe that our dataset can strongly contribute to the controllability of image generation methods in creative industry domains.
    \item For a more effective metric, we provide multiple reference images in the test set and each reference image has fine-grained scores.
    \item We apply two state-of-the-art models to our dataset and then find some valuable phenomena that we believe can guide the direction of subsequent research.
\end{itemize}

\section{Related work}

\paragraph{Datasets.} 
\textbf{a) Datasets with neither prompt nor sketch}.
The CIFAR-10 \cite{abouelnaga2016cifar} dataset consists of 60000 32x32 color images in 10 classes, with 6000 images per class. 
ImageNet \cite{deng2009imagenet} is an image dataset organized according to the WordNet \cite{miller1995wordnet} hierarchy, with 14 million images and 21841 synsets. 
\textbf{b) The dataset with prompt only}. 
CC \cite{sharma2018conceptual} present a dataset of image caption annotations, which contains 3 million images and represents a wider variety of both images and image caption styles. 
CC12M \cite{changpinyo2021conceptual} introduce a vision-and-language pre-training resource obtained by leveraging noisy Web-scale image-text pairs. 
COYO-700M \cite{kakaobrain2022coyo-700m} collecting many informative pairs of alt-text and its associated image in HTML documents which contains 747M image-text pairs. 
LAION-400M \cite{schuhmann2021laion} and LAION-5B \cite{schuhmann2022laion} are datasets with CLIP-filtered 400 million and 5 billion image-text pairs respectively, their CLIP embeddings and kNN indices that allow efficient similarity search. 
Danbooru2021 \cite{danbooru2021} is a large-scale anime image database with 4.9m+ images annotated with 162m+ tags.
\textbf{c) The dataset with sketch only}. 
Photo-Sketching \cite{li2019photo} use a crowdsourcing platform to collect 5000 high-quality drawings on outdoor images. 
Sketchy \cite{sangkloy2016sketchy} ask crowd workers to sketch particular photographic objects sampled from 125 categories and acquire 75471 sketches of 12500 objects. 
\textbf{d) The dataset with both prompt and sketch}. 
Multi-Modal CelebA-HQ \cite{karras2017progressive} is a large-scale face image dataset that has 30,000 high-resolution face images, each having a high-quality segmentation mask, sketch, and descriptive text. 
M2C-Fashion \cite{zhang2021m6} is a multi-modal large-scale clothing dataset.

\paragraph{Prompt.}
Danbooru2021 \cite{branwen2019danbooru2019} is an anime image database with 4.9 million images annotated with 162 million tags which associate several tags to one image.
CC12M \cite{changpinyo2021conceptual} is a dataset with 12 million image-text pairs which provide one caption to one image.
MS-COCO \cite{lin2014microsoft} contains 328000 images which provide five written caption descriptions for each image. 

\paragraph{Sketch.} 
Boundary Detection \cite{geman1990boundary} is a joint probability distribution for the array of pixel gray levels and an array of labels.
Photo-Sketching \cite{li2019photo} collect a new dataset of contour drawings and propose a learning-based method that resolves diversity in the annotation and, unlike boundary detectors, can work with imperfect alignment of the annotation and the actual ground truth.
SketchyScene \cite{zou2018sketchyscene} is created through a novel and carefully designed crowdsourcing pipeline, enabling users to efficiently generate large quantities of realistic and diverse scene sketches. SketchyScene contains more than 29,000 scene-level sketches, 7,000+ pairs of scene templates and photos, and 11,000+ object sketches.

\paragraph{Image Generation methods}
\textbf{a) Support neither prompt nor sketch.}
StyleGAN \cite{karras2019style} propose an alternative generator architecture for generative adversarial networks, borrowing from style transfer literature. 
CycleGAN \cite{zhu2017unpaired} present an approach for learning to translate an image from a source domain X to a target domain Y in the absence of paired examples. 
\textbf{b) Support only prompt.}
Make-A-Scene \cite{gafni2022make} propose a novel text-to-image method that enables a simple control mechanism complementary to text in the form of a scene.
DALL·E \cite{ramesh2021zero} based on a transformer that autoregressively models the text and image tokens as a single stream of data. 
\textbf{c) Support only sketch.}
SketchyGAN \cite{chen2018sketchygan} propose a novel Generative Adversarial Network (GAN) approach that synthesizes plausible images from 50 categories including motorcycles, horses, and couches.
Interactive Sketch \cite{ghosh2019interactive} propose an interactive GAN-based sketch-to-image translation method that helps novice users easily create images of simple objects. 
\textbf{d) Support both prompt and sketch.}
PoE-GAN \cite{huang2022multimodal} is a product-of-experts generative adversarial networks framework, which can synthesize images conditioned on multiple input modalities or any subset of them, even the empty set. 
M6-UFC \cite{zhang2021m6} propose a new two-stage architecture to unify any number of multi-modal controls, in which both the diverse control signals and the synthesized image are uniformly represented as a sequence of discrete tokens to be processed by Transformer.
Stable Diffusion (SD) \cite{rombach2022high} apply diffusion model training in the latent space of powerful pretrained autoencoders and achieve a new state of the art for image inpainting and highly competitive performance on various tasks. NovelAI \cite{NovelAI} is a production focused on anime art generation based on stable diffusion method. 

\paragraph{Evaluation Methods.}
Inception Score \cite{salimans2016improved} (IS) evaluates GANs by using the KL divergence between properties captured by a pre-trained network (InceptionNet \cite{szegedy2016rethinking}, trained on the ImageNet \cite{deng2009imagenet} dataset).
Fréchet Inception Distance (FID) \cite{heusel2017gans} evaluates the generated data and real data by the Fréchet distance between two continuous multivariate Gaussian distributions constructed from the mean and variance of the feature of InceptionNet.

\section{Image Collection}
In order to build a dataset for the creative industry domain, we need to solve the following problems: a) What is the creative industry domain? b) How to get the images belonging to the creative industry domain? c) How to get prompts and sketches?

\begin{table*}[t!]
\centering
\caption{The classification of the creative industry.}
\label{industry_task_categories}
\begin{tabular}{ll}\toprule[2pt]
Name & Describe\\
\midrule[1pt]
concept design & game concept design, film concept design, cartoon animation, illustration, costume\\
graphic design & posters, packaging, product design, publications\\
3D-CG & 3D modeling, 3D animation, visual effects design, CG environment, renderings\\
outdoor design & architecture, garden, landscape, outdoor environment, sculpture\\
\bottomrule[1pt]
\end{tabular}  
\end{table*}

\begin{figure*}[t!]
\centering
\includegraphics[scale=0.22]{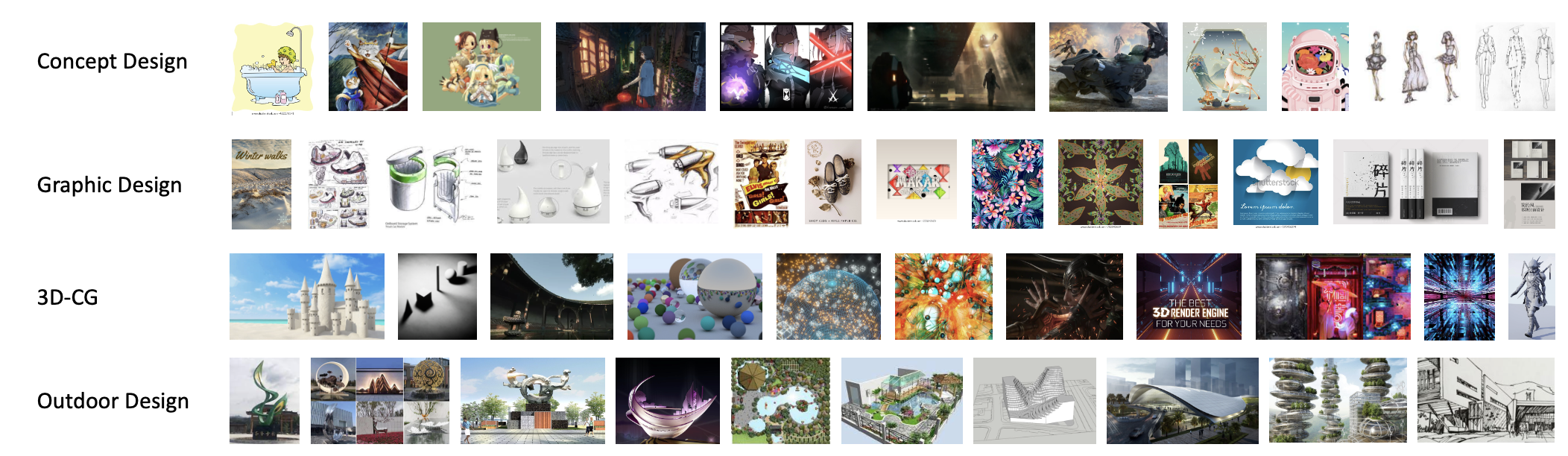}
\centering
\caption{Examples of images in each category in the creative industry domain.
\label{sample_category}}
\end{figure*}

\subsection{Creative Industry Domain}
\label{sec:industry_domain}
Due to the lack of a comprehensive and credible classification system for the creative industrial domain, we divide the creative industry domain into 4 categories by the suggestion made by an art professional who is inspired by the classification system of art-pro websites \cite{artstation, pinterest}. As shown in Table \ref{industry_task_categories} and Fig. \ref{sample_category}, the 4 categories are Concept Design (CD), Graphic Design (GD), 3D-CG, and Outdoor Design (OD). 

\subsection{Image Source}
Since a large number of datasets with prompt already exist, we directly select the images belonging to the creative industry domain from the existing datasets CC12M \cite{changpinyo2021conceptual} and Danbooru2021
\cite{danbooru2021}, thus reusing the prompts from the previous datasets. Since the images in Danbooru2021 have only tags, we splice the tags together to construct a sentence as its prompt.

\subsection{Image Selection}
We only select images that match the creative industry domain by manual annotation. Images that do not belong to these categories will not be selected. We employ 5 annotators and train them until the accuracy of their annotation is higher than 99.0\%. The distribution of categories is shown in Table \ref{Selected_image_category_distribution}. Each image in the dataset is labeled with the category it belongs to.

\begin{table}
\centering
\caption{Selected image category distribution.}
\label{Selected_image_category_distribution}
\begin{tabular}{lrrrr}\toprule[2pt]
label & CD & GD & 3D-CG & OD\\
\midrule[1pt]
Percent & 42.5 & 33.9 & 17.5 & 6.1\\
\bottomrule[1pt]
\end{tabular}  
\end{table}

\section{Prompt and Sketch}
We need to give each image a prompt and a sketch. As the images were selected from CC12M and Danbooru2021, they all have a prompt. Therefore, we only need to generate a sketch for each image. Considering that some art creators sketch with just a few strokes, while some art creators sketch in great detail, we should provide sketches made up of different numbers of strokes.

\subsection{Sketch Annotation Rules}
To assess the quality of the sketch, the following two evaluation dimensions were developed: shape fidelity and line quality (cf. Appendix E). Note that whether a sketch has only a few strokes or a lot of strokes does not affect the evaluation of the quality of the sketch, i.e. the number of strokes in the sketch does not matter. This principle ensures that our dataset does not prefer simple sketches or prefer complex sketches. Figure \ref{test_data_example} shows examples of sketches with their evaluation score.
\begin{figure}[t!]
\centering
\includegraphics[scale=0.17]{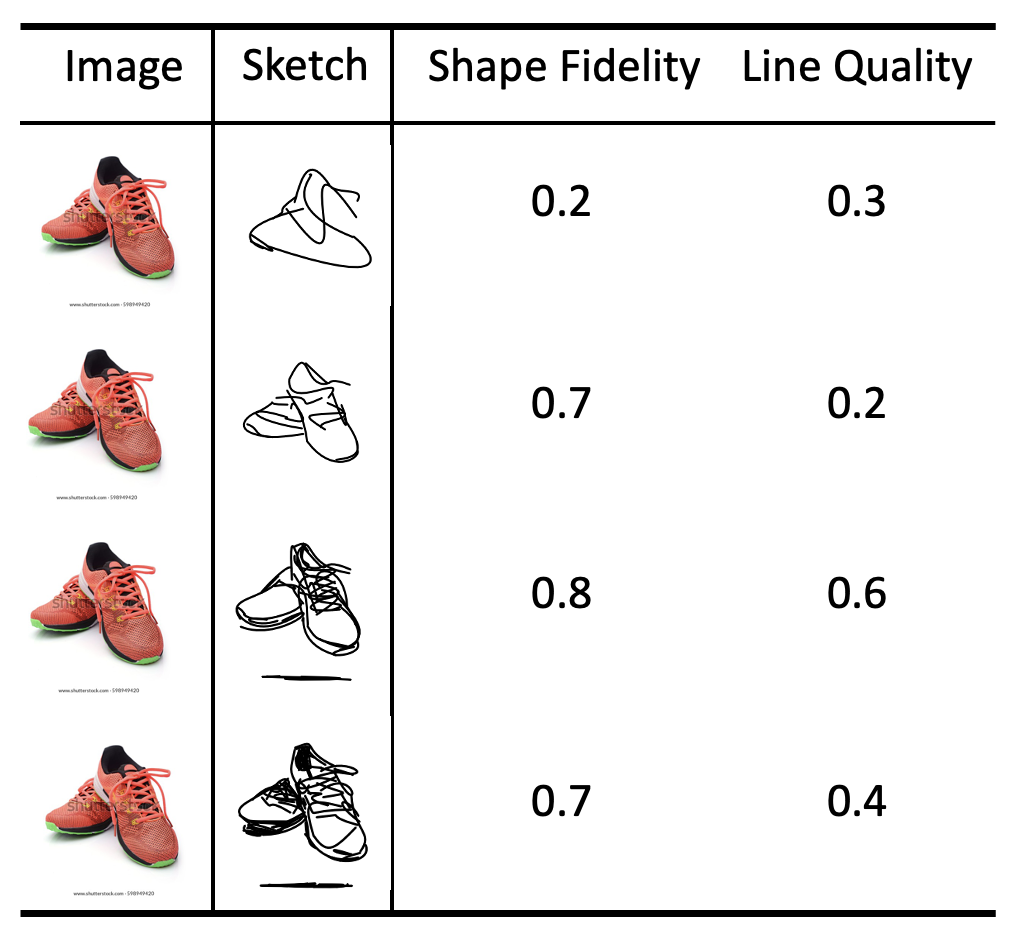}
\centering
\caption{Examples of sketches with their evaluation score.
\label{sketch_rule_case}}
\end{figure}

\subsection{Method of Obtain Sketch}
\begin{figure}[t!]
\centering
\includegraphics[scale=0.37]{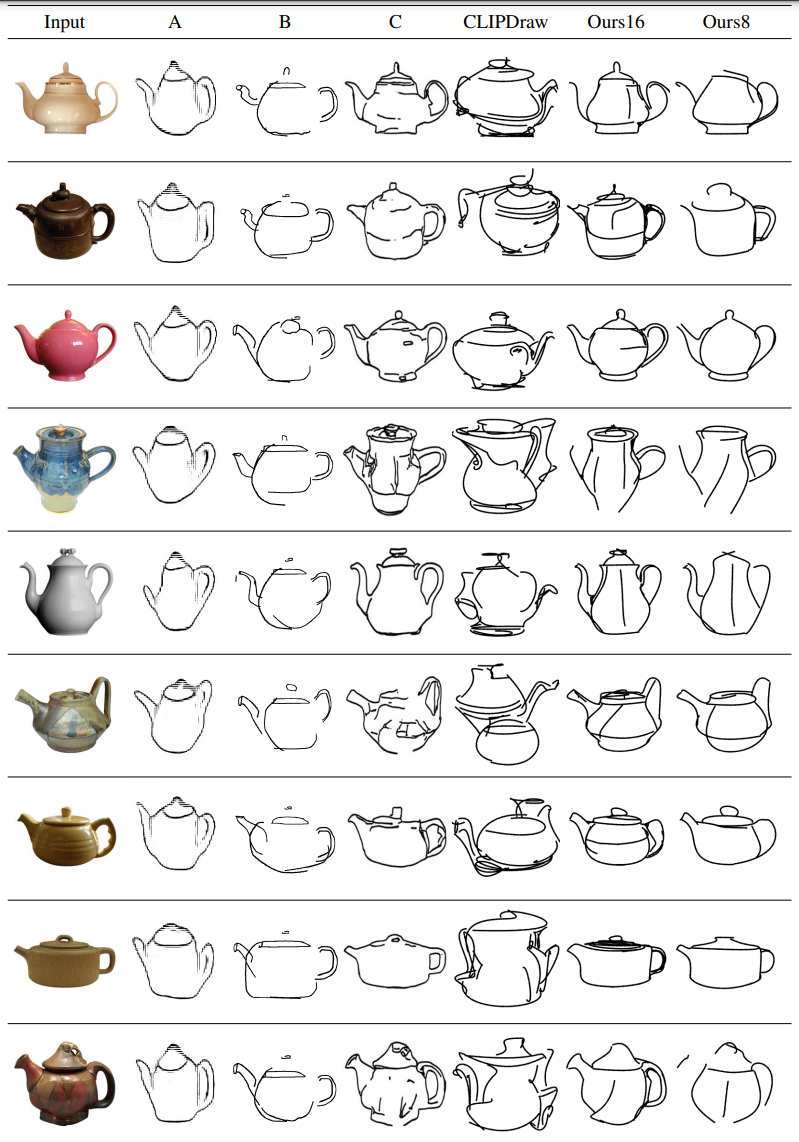}
\centering
\caption{Comparison to competitor methods. Left to right : (A) Kampelmuhler and Pinz \cite{kampelmuhlerP2020synthesizing}, (B) Li et al. \cite{li2017free}, (C) Li et al. \cite{li2019photo}, and CLIPDraw \cite{frans2106clipdraw}. Our16 and Our8 refer to the sketches generated by CLIPasso model with strokes of 16 and 8 respectively.
\label{sketch_compare_pic}}
\end{figure}

Due to the high cost of sketch annotation, we use a model named CLIPasso \cite{vinker2022clipasso} to obtain sketches at a low cost. 
It has been proven that method CLIPasso is better than competing methods such as BCDE (A) Kampelmuhler and Pinz \cite{kampelmuhlerP2020synthesizing} , (B) Li et al. \cite{li2017free}, (C) Li et al. \cite{li2019photo}, and CLIPDraw \cite{frans2106clipdraw}. And a case study is shown in Fig.\ref{sketch_compare_pic}. \cite{vinker2022clipasso}

\begin{figure}[t!]
\centering
\includegraphics[scale=0.6]{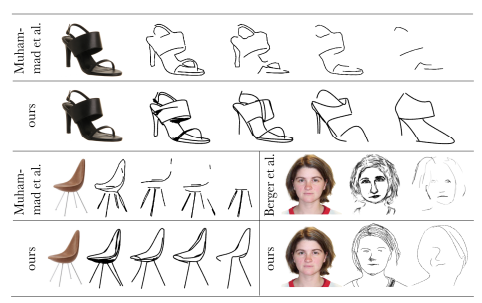}
\centering
\caption{Levels of Abstraction Comparison. In the top and left parts are comparisons to Muhammad et al. \cite{muhammad2018learning} and in the right bottom part is a comparison to Berger et al. \cite{berger2013style}. The leftmost column shows the input image, and the next four columns show different levels of abstraction.
\label{sketch_score_pic_2}}
\end{figure}

Furthermore, as examples shown in Fig.\ref{sketch_score_pic_2}, CLIPasso is proven superior on generate sketches with different levels of abstraction. See Appendix H for details of the parameter setting of CLIPasso.

\subsection{Sketch selection}
The quality of the sketches generated by CLIPasso varies. In order to improve the quality of sketches in our dataset, we asked the annotators to evaluate all sketches according to sketch annotation rules and remove all sketches with shape fidelity less than 0.5 or line quality less than 0.2.

\section{Reference Images and Fine-Grained Scores}
Since our dataset focuses on the prompt and the sketch, we wonder whether the generated results satisfy the requirements of the prompt and the sketch. The previous evaluation methods cannot meet our needs, because they have two shortcomings: a) There is no standard answer for the image generation task, and using only one golden reference image limits the accuracy of the evaluation. b) It is difficult to assess the similarity between the image and prompt or sketch. 
To alleviate these problems, we enhance the test set with multiple reference images and fine-grained scores. And then we propose a new evaluation method.

\subsection{Generate Reference Images}
We use two state-of-the-art image generation models SD \cite{rombach2022high} and NovelAI \cite{NovelAI} that support both prompt and sketch to generate candidate reference images. For detail, we input prompt and sketch to the two models and ask each model to generate 10 images. We call these 20 generated images candidate reference images. See Appendix G for details of the parameter setting of these models.

\subsection{Reference Image Annotation Rules}
\label{sec:evaluation_metrics}
To annotate the level of consistency of a reference image and a condition consisting of one prompt and one sketch, fine-grained image annotation rules (cf. Appendix F) are designed by an art professional (cf. Appendix A). As shown in Fig.\ref{test_data_example}, the fine-grained score contains PSC, SCS, VES, and CS.

\paragraph{Data Split.}
We randomly selected 500 images from the training set as the test set and the rest images as the Train set.

\paragraph{Image Selection.}
We obtain 20 generated image candidate reference images for each image in the test set. Then we annotate all the candidate reference images and delete the images of which the Composite Score (CS) is lower than 0.7.

\begin{figure}[t!]
\centering
\includegraphics[scale=0.17]{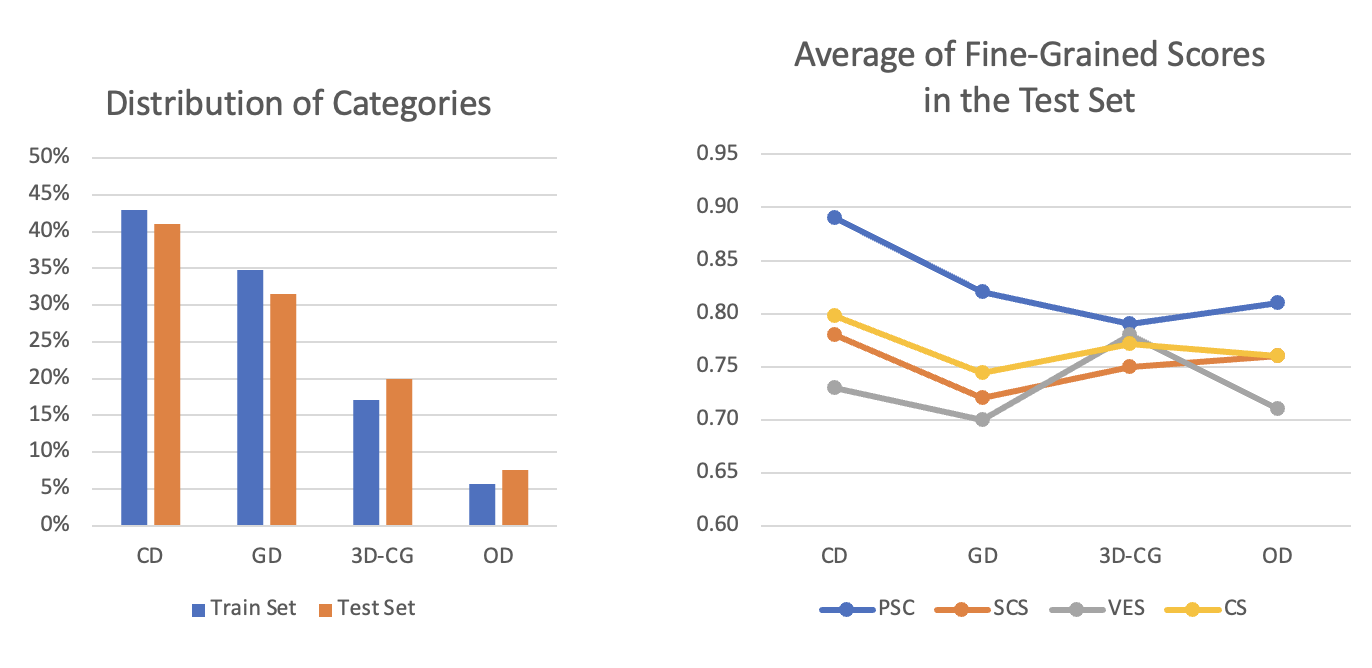}
\centering
\caption{Statistic of our dataset.
\label{statistic_our_dataset}}
\end{figure}

\paragraph{Statistic.}
Some statistic of our dataset is shown in Fig.\ref{statistic_our_dataset}.

\subsection{Weighted-FID}
\label{sec:weighted_fid}
Based on the information from our evaluation dataset, we propose a new evaluation method named weighted-FID (wFID). The calculation process of wFID consists of 2 steps: 1) Using FID to calculate the scores of the image to be evaluated with each reference image. 2) The CS of each reference image is considered as the weight to calculate a weighted average score, i.e., wFID. To test the wFID method we make an experiment, as shown in Table \ref{auto_metric}: a) We randomly selected 10 test cases from the test set. b) We used NovalAI to generate an image for each case, and the annotators scored it to get the manual score. c) We used the golden reference image to calculate the FID score, and all reference images to calculate the wFID score. d) We sorted the 10 generated images by each evaluation method and found that the agreement rate between manual score and wFID score is higher than that of FID score.

ATTENTION: the wFID is only one of the many possible ways to utilize our test set. We believe that the value of Multiple Reference Images and Fine-Graied scores has not been fully exploited.
\begin{table}
\centering
\caption{An comparison of human score, FID score, and wFID score.}
\label{auto_metric}
\begin{tabular}{lrrrrrrrrrr}\toprule[2pt]
Case & C &G & B &F & A & H& E & J & D & I\\
\midrule[1pt]
Human & 1 & 2 & 3 & 4 & 5 & 6 & 7 & 8 & 9 & 10\\
FID & 6 & 10 & 7 & 9 & 2 & 4 & 6 & 1 & 3 & 8\\
wFID & 2 & 4 & 1 & 3 & 10 & 6 & 8 & 7 & 5 & 9\\
\bottomrule[1pt]
\end{tabular}  
\end{table}

\section{Discussion}
\label{sec:Discussion}
We found some phenomena of the current SOTA model on our dataset. Note that we did not fine-tune these models.

\subsection{Prompts and Sketches is Effective.}
As shown in Fig. \ref{prompt_and_sketch}, we find that both prompt and sketch can control the images generated by models, but images generated by diffusion models have a big gap in prompt consistency compared to the manually selected images. The model currently does not combine the information of prompt and sketch well to generate images, which needs to be improved and is a promising research direction. And the prompt is more highly valued than the sketch in both models.

\subsection{The difference among four industry task categories}
The images generated by the model \cite{rombach2022high, NovelAI} were analyzed in 4 Industry Task categories scenarios. In the Concept design task, as shown in Fig. \ref{sample_category}, the images are mostly in the form of cartoons and animation. This task does not require high details of characters and things in the images. After manual evaluation, the images generated by diffusion models often have a higher CS. As shown in Fig. \ref{sample_category}, the Graphic design task compared with the Concept design task requires more details, but most of the images generated by diffusion models are not fine enough, so there are fewer acceptable images. In the 3D-CG task, the objects in the images are not required to conform to common sense logic. For example, a prompt like ``a lamp without a light bulb'' is acceptable, and the images generated by diffusion models are of low quality. We analyzed this problem in \ref{discussion3}. In the Outdoor design task, the prompt is mostly a text describing the whole. For example, ``a snow-capped mountain.''. The content of the prompt is not much, and the generated images are more likely to conform to the prompt but are missing a lot of detail compared to the original image. So the images generated by diffusion models have a higher PCS and a lower SCS.


\subsection{The competition between prompt and sketch}
We analyzed whether there is a competitive relationship between prompt and sketch. We conducted a case study for the following two queries.

1. when the prompt is rich, do the images generated by diffusion models follow the prompt more than the line shape of the sketch?

2. When there is less or no content in the prompt, do the images generated by diffusion models follow the line shape of the sketch more than the prompt?

\begin{figure*}[t!]
\centering
\includegraphics[scale=0.4]{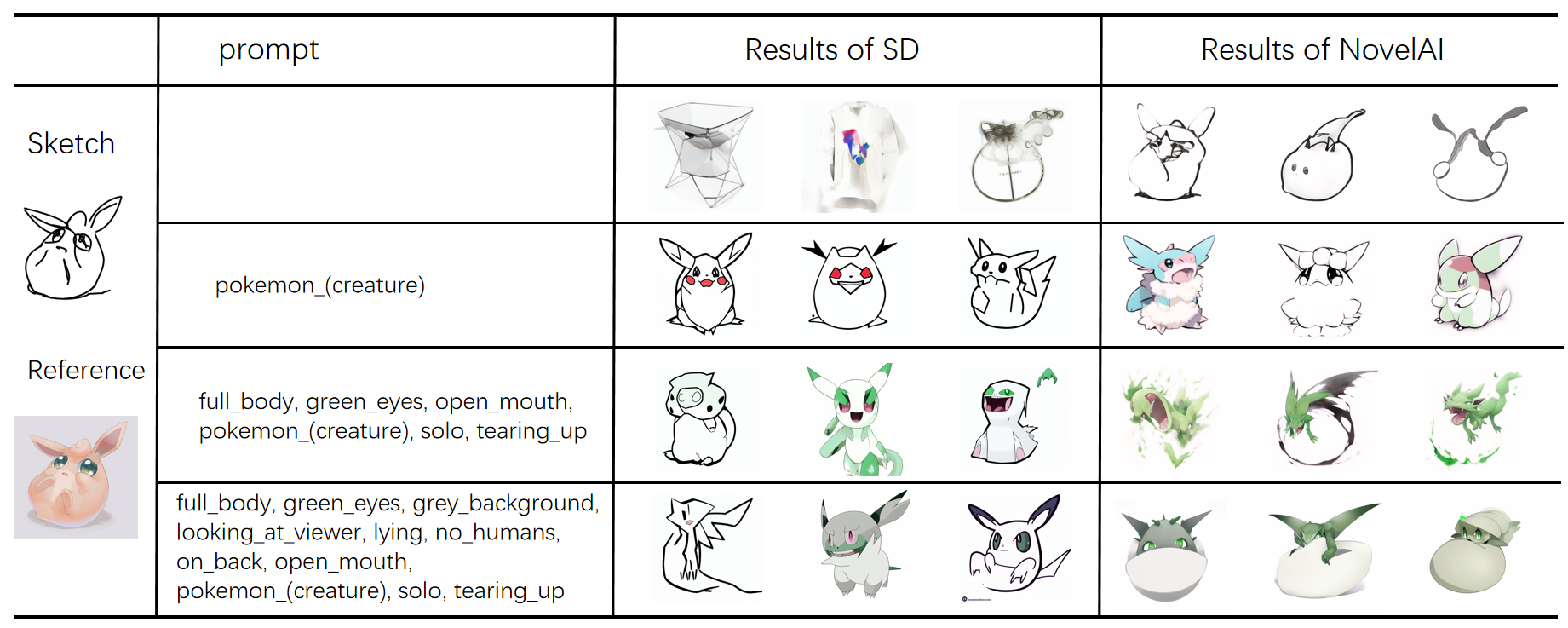}
\centering
\caption{The case study of the competition between prompt and sketch
\label{The competition between prompt and sketch}}
\end{figure*}

As shown in Fig. \ref{The competition between prompt and sketch}, the richness of the prompt has nothing to do with whether the images generated by diffusion models follow the line shape of the sketch more or not, or whether they follow the prompt more.

\subsection{The analysis of anti-common sense data \label{discussion3}}
\begin{figure}[t!]
\centering
\includegraphics[scale=0.25]{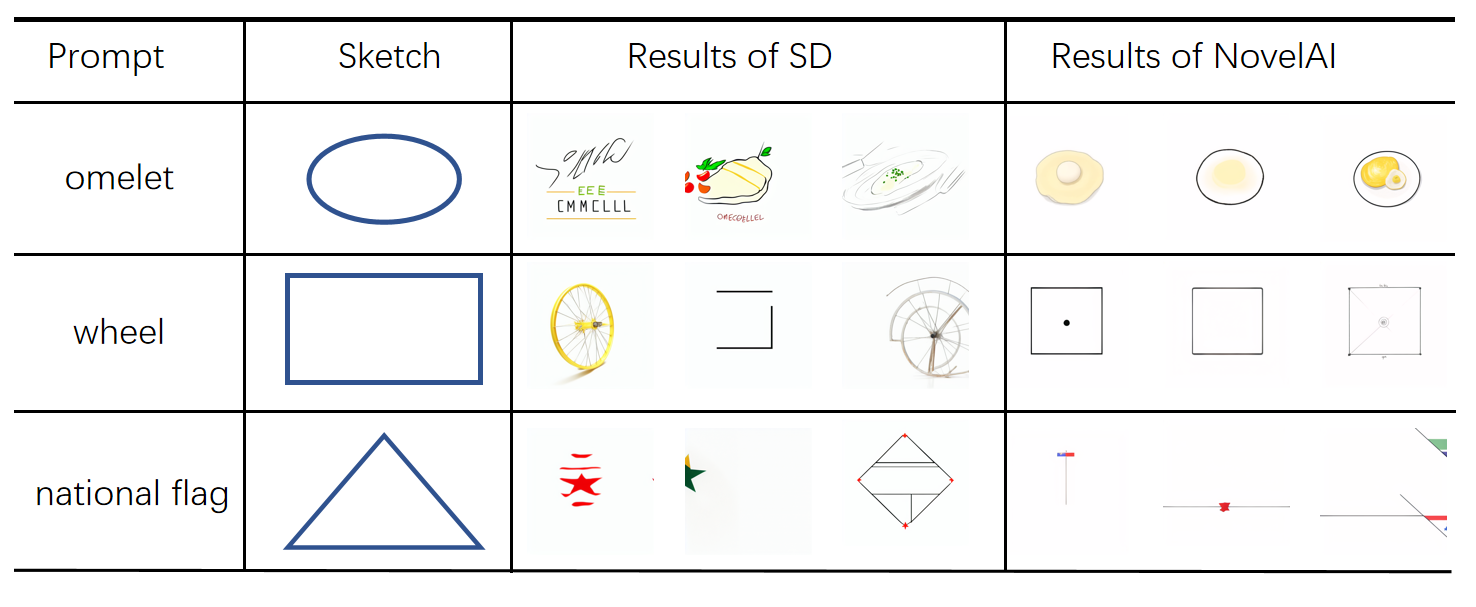}
\centering
\caption{Examples of anti-common sense data
\label{omelet}}
\end{figure}

As shown in Fig. \ref{prompt_and_sketch}, in the case where the prompt is ``ball'', the PCS of the images generated by diffusion models \cite{rombach2022high, NovelAI} is significantly lower than that of the images generated by diffusion models when the input sketch is an ellipse than when the input sketch is a circle. The same phenomenon is observed in the case where the prompt is ``fruit''. We deduce that this phenomenon is caused by the data distribution, which means that most of the fruits and balls in the training set of these models are round, with few oval fruits or oval balls. To verify this conclusion, as shown in Fig. \ref{omelet}, we keep the oval sketch unchanged and changed the prompt to ``omelet''. After manual evaluation, diffusion models generate images with a mean PCS value greater than 0.8. Further, we also generate images when the prompt is ``wheel'' with the input sketch as a square and the prompt is ``national flag'' with the input sketch as a triangle. After manual evaluation, their mean CS is less than 0.2 and 0.1, respectively. It is reasonable to assume that uncommon combinations of the prompt and the image in the training data will not perform well when inferred.

\subsection{Semantic confusion}
\begin{figure}[t!]
\centering
\includegraphics[scale=0.25]{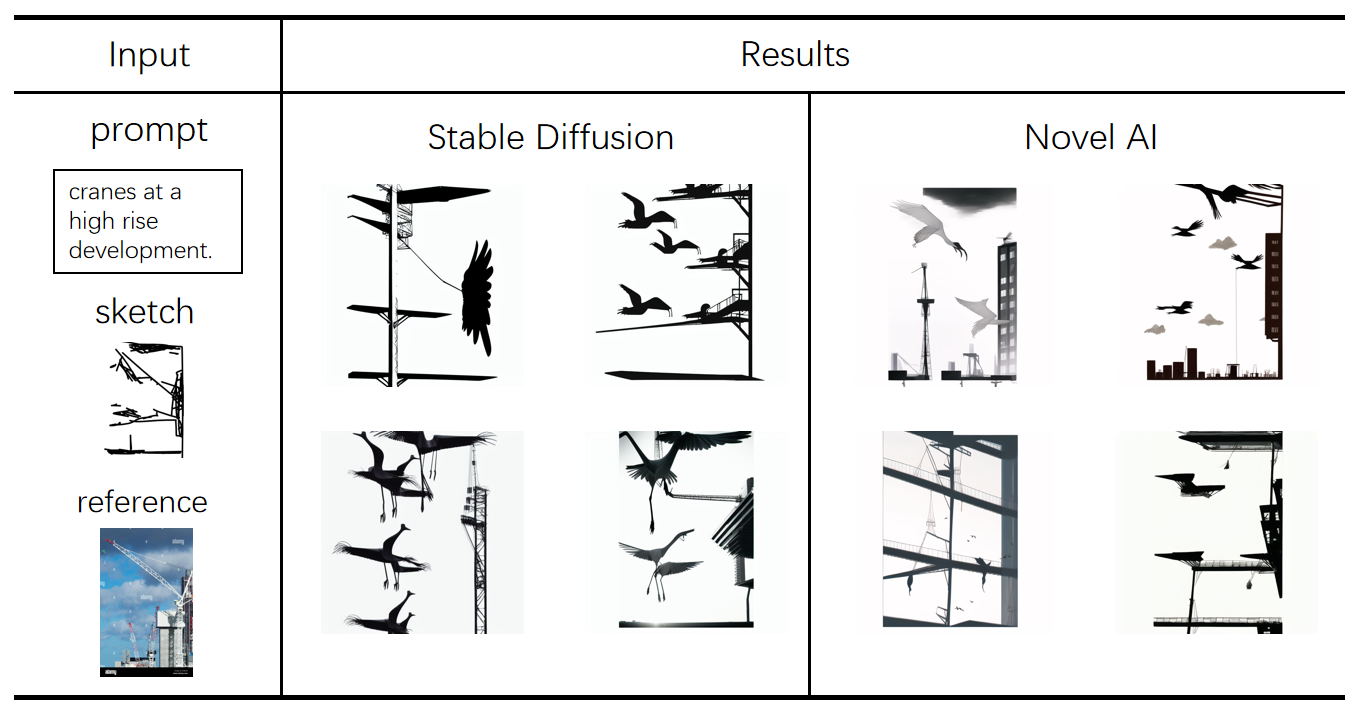}
\centering
\caption{The animal ``crane'' and the machine ``crane'' both appear in the images
\label{Semantic confusion}}
\end{figure}

When a word appears polysemy in a text, the diffusion models \cite{rombach2022high, NovelAI} can not disambiguate the text semantics well. As shown in Fig. \ref{Semantic confusion}, ``Carnes'' is a polysemous word in the prompt ``cranes at a high rise development.''. 

\subsection{The influence of sketch drawing form}
\begin{figure}[t!]
\centering
\includegraphics[scale=0.25]{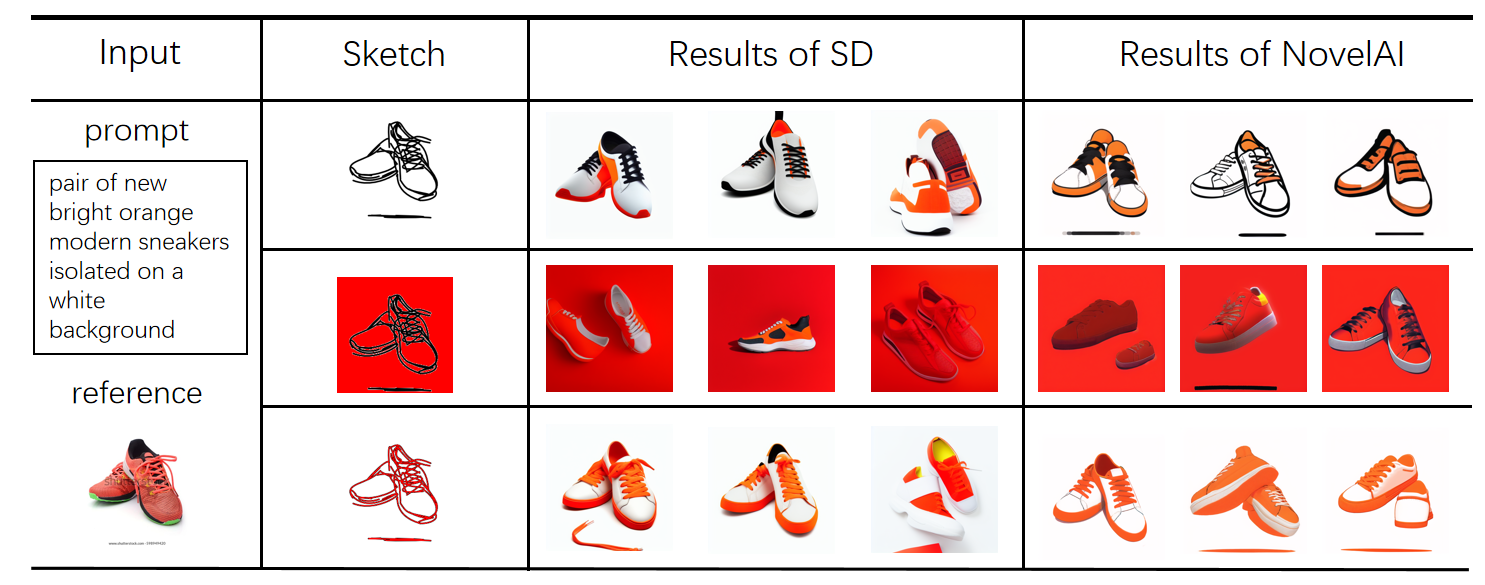}
\centering
\caption{Examples of three drawing forms
\label{The influence of sketch drawing form}}
\end{figure}

As shown in Fig. \ref{The competition between prompt and sketch} \ref{omelet} \ref{Semantic confusion} \ref{The influence of cfg scale parameter} \ref{Realistic images}, the sketches are in the drawing form of a black line on white background, the images generated by diffusion models \cite{rombach2022high, NovelAI} are also often in the form of black lines on a white background. In order to investigate whether this form of painting leads to a specific style of painting in the images generated by the model, we used the PS tool to modify the black lines and the white background to red respectively. As shown in Fig. \ref{The influence of sketch drawing form}, We found that the drawing form of the sketch does have a great impact on the generated image. Take the second line of the image as an example, in the case that the background of the sketch is red, there is no way to generate an image with a white background even if the prompt contains a description that modifies the background to white. Moreover, when the line color or background color of the sketch is changed, the SC of the generated images is relatively lower.

\subsection{The influence of cfg scale parameter}
\begin{figure}[t!]
\centering
\includegraphics[scale=0.25]{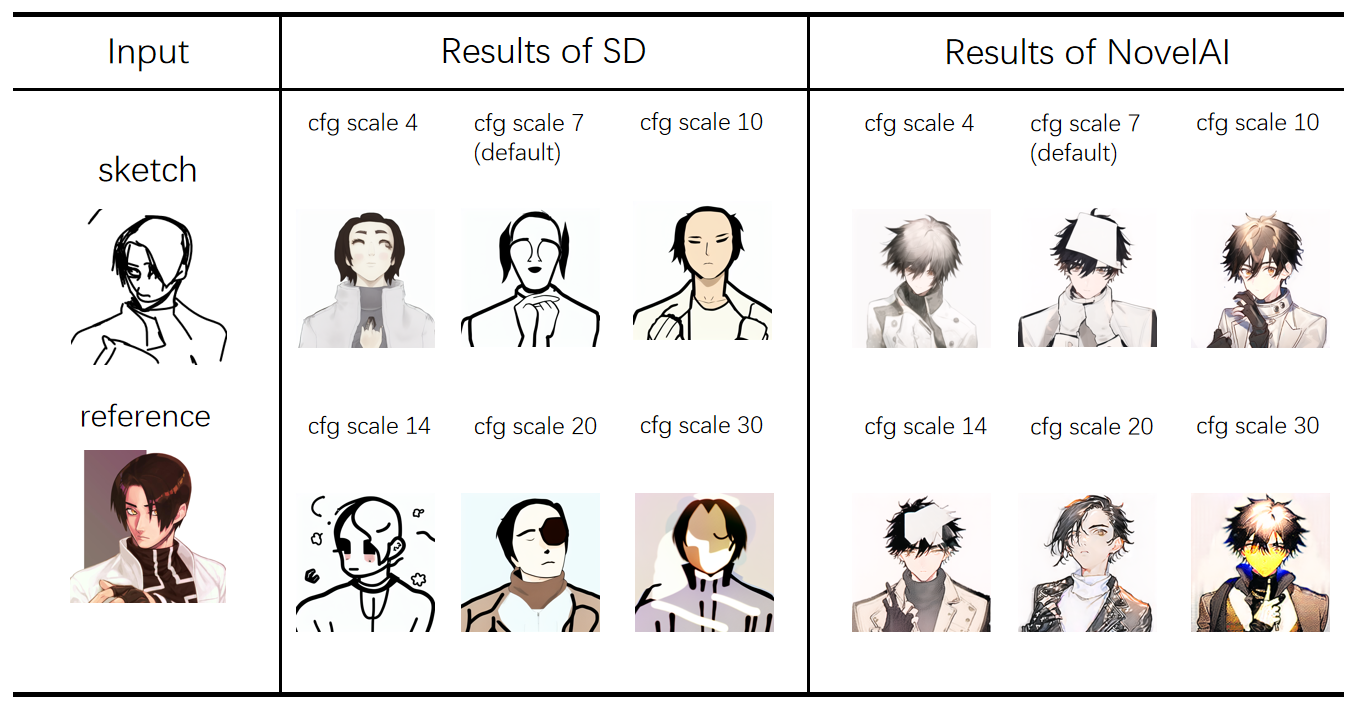}
\centering
\caption{Prompt is ``1boy, black hair, brown eyes, brown hair, cross print, curtained hair, fingerless gloves, gloves, jacket, looking at viewer, male focus, parted hair, solo, turtleneck, upper body, white jacket''
\label{The influence of cfg scale parameter}}
\end{figure}

Stable-diffusion-webui, a tool that generates the images, provides cfg scale parameter, its full name is Classifier Free Guidance Scale, which means how strongly the image should conform to prompt - lower values produce more creative results. \cite{webui} When generating images, the cfg scale parameter can be used to adjust the faithfulness. As shown in Fig. \ref{The influence of cfg scale parameter}, this parameter does not make the pictures generated by diffusion models more faithful to the lines of the sketch. Even, too large cfg scale parameter will cause diffusion models to generate bad images.

\subsection{The comparison of two diffusion models}
\begin{figure}[t!]
\centering
\includegraphics[scale=0.25]{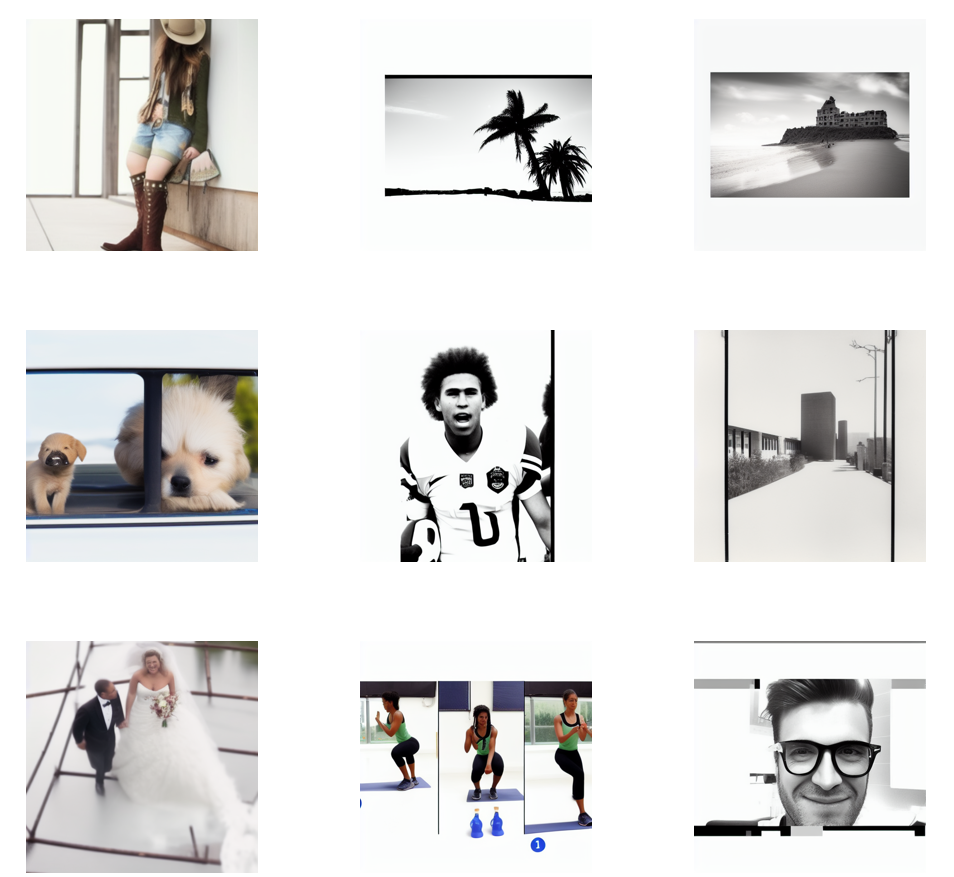}
\centering
\caption{Examples of realistic images generated by stable diffusion v1.4
\label{Realistic images}}
\end{figure}

The two diffusion models used in our experiment, stable diffusion v1.4 \cite{rombach2022high} can generate more realistic pictures, and the images generated by novel AI \cite{NovelAI} are basically in the form of animation and cartoon. Based on the test set data we provided, the SD model generated realistic images with roughly 9\% probability, while the novel AI generated realistic images with roughly 2\% probability. As shown in Fig. \ref{Realistic images}, We show several realistic images generated by stable diffusion v1.4. Besides, As shown in Fig. \ref{The competition between prompt and sketch} \ref{The influence of cfg scale parameter}, It is worth noting that novel AI is more suitable for the Concept design task because of the high quality of the generated cartoon and anime images.

\section{Appendix Overview}
\paragraph{A) Train set examples.} Show several cases in train set.

\paragraph{B) Test set examples.} Show several cases in test set.

\paragraph{C) Educational background of annotators.} Detailed information on the educational background of each marker.
\paragraph{D) Training program for annotators.} To ensure annotation accuracy of annotators.
\paragraph{E) Scoring criteria of sketch.} Details the criteria used by annotators to assess the quality of Sketch.
\paragraph{F) Scoring criteria of reference.} Details the criteria used by annotators to assess the quality of Sketch.
\paragraph{G) Setting of SD and Noval AI.} The setting of SD and Noval AI when generating candidate reference images.
\paragraph{H) Setting of CLIPasso.} The setting of CLIPasso when generating sketches.

{\small
\bibliographystyle{ieee_fullname}
\bibliography{egbib}

\begin{thebibliography}{10}\itemsep=-1pt

\bibitem{webui}
A browser interface based on gradio library for stable diffusion.
\newblock https://github.com/AUTOMATIC1111/stable-diffusion-webui, 2022.

\bibitem{abouelnaga2016cifar}
Yehya Abouelnaga, Ola~S Ali, Hager Rady, and Mohamed Moustafa.
\newblock Cifar-10: Knn-based ensemble of classifiers.
\newblock In {\em 2016 International Conference on Computational Science and
  Computational Intelligence (CSCI)}, pages 1192--1195. IEEE, 2016.

\bibitem{anantrasirichai2021artificial}
Nantheera Anantrasirichai and David Bull.
\newblock Artificial intelligence in the creative industries: a review.
\newblock {\em Artificial Intelligence Review}, pages 1--68, 2021.

\bibitem{NovelAI}
Anlatan.
\newblock Novalai: Driven by ai, painlessly construct unique stories, thrilling
  tales, seductive romances, or just fool around. anything goes!, 2022.

\bibitem{danbooru2021}
Anonymous, Danbooru community, and Gwern Branwen.
\newblock Danbooru2021: A large-scale crowdsourced and tagged anime
  illustration dataset.
\newblock \url{https://www.gwern.net/Danbooru2021}, January 2022.
\newblock Accessed: DATE.

\bibitem{banks2009after}
Mark Banks and Justin O’Connor.
\newblock After the creative industries.
\newblock 2009.

\bibitem{berger2013style}
Itamar Berger, Ariel Shamir, Moshe Mahler, Elizabeth~J. Carter, and Jessica~K.
  Hodgins.
\newblock Style and abstraction in portrait sketching.
\newblock {\em {ACM} Trans. Graph.}, 32(4):55:1--55:12, 2013.

\bibitem{branwen2019danbooru2019}
Gwern Branwen and A Gokaslan.
\newblock Danbooru2019: A large-scale crowdsourced and tagged anime
  illustration dataset, 2019.

\bibitem{kakaobrain2022coyo-700m}
Minwoo Byeon, Beomhee Park, Haecheon Kim, Sungjun Lee, Woonhyuk Baek, and
  Saehoon Kim.
\newblock Coyo-700m: Image-text pair dataset.
\newblock \url{https://github.com/kakaobrain/coyo-dataset}, 2022.

\bibitem{changpinyo2021conceptual}
Soravit Changpinyo, Piyush Sharma, Nan Ding, and Radu Soricut.
\newblock Conceptual 12m: Pushing web-scale image-text pre-training to
  recognize long-tail visual concepts.
\newblock In {\em Proceedings of the IEEE/CVF Conference on Computer Vision and
  Pattern Recognition}, pages 3558--3568, 2021.

\bibitem{chen2020deepfacedrawing}
Shu-Yu Chen, Wanchao Su, Lin Gao, Shihong Xia, and Hongbo Fu.
\newblock Deepfacedrawing: Deep generation of face images from sketches.
\newblock {\em ACM Transactions on Graphics (TOG)}, 39(4):72--1, 2020.

\bibitem{chen2018sketchygan}
Wengling Chen and James Hays.
\newblock Sketchygan: Towards diverse and realistic sketch to image synthesis.
\newblock In {\em Proceedings of the IEEE Conference on Computer Vision and
  Pattern Recognition}, pages 9416--9425, 2018.

\bibitem{deng2009imagenet}
Jia Deng, Wei Dong, Richard Socher, Li-Jia Li, Kai Li, and Li Fei-Fei.
\newblock Imagenet: A large-scale hierarchical image database.
\newblock In {\em 2009 IEEE conference on computer vision and pattern
  recognition}, pages 248--255. Ieee, 2009.

\bibitem{ekwaro2016uncertainty}
Stephen Ekwaro-Osire, Ricardo Cruz-Lozano, Haileyesus~B Endeshaw, and
  Jo{\~a}o~Paulo Dias.
\newblock Uncertainty in communication with a sketch.
\newblock {\em Journal of Integrated Design and Process Science}, 20(4):43--60,
  2016.

\bibitem{artstation}
Inc. Epic~Games.
\newblock Artstation: provides an awesome browsing and discovery experience,
  enabling you to view thousands of artworks by the world's best artists.,
  2022.

\bibitem{frans2106clipdraw}
Kevin Frans, Lisa~B. Soros, and Olaf Witkowski.
\newblock Clipdraw: Exploring text-to-drawing synthesis through language-image
  encoders.
\newblock {\em CoRR}, abs/2106.14843, 2021.

\bibitem{gafni2022make}
Oran Gafni, Adam Polyak, Oron Ashual, Shelly Sheynin, Devi Parikh, and Yaniv
  Taigman.
\newblock Make-a-scene: Scene-based text-to-image generation with human priors.
\newblock {\em arXiv preprint arXiv:2203.13131}, 2022.

\bibitem{geman1990boundary}
Donald Geman, Stuart Geman, Christine Graffigne, and Ping Dong.
\newblock Boundary detection by constrained optimization.
\newblock {\em IEEE Transactions on pattern analysis and machine intelligence},
  12(7):609--628, 1990.

\bibitem{ghosh2019interactive}
Arnab Ghosh, Richard Zhang, Puneet~K Dokania, Oliver Wang, Alexei~A Efros,
  Philip~HS Torr, and Eli Shechtman.
\newblock Interactive sketch \& fill: Multiclass sketch-to-image translation.
\newblock In {\em Proceedings of the IEEE/CVF International Conference on
  Computer Vision}, pages 1171--1180, 2019.

\bibitem{goodfellow2020generative}
Ian Goodfellow, Jean Pouget-Abadie, Mehdi Mirza, Bing Xu, David Warde-Farley,
  Sherjil Ozair, Aaron Courville, and Yoshua Bengio.
\newblock Generative adversarial networks.
\newblock {\em Communications of the ACM}, 63(11):139--144, 2020.

\bibitem{heusel2017gans}
Martin Heusel, Hubert Ramsauer, Thomas Unterthiner, Bernhard Nessler, and Sepp
  Hochreiter.
\newblock Gans trained by a two time-scale update rule converge to a local nash
  equilibrium.
\newblock {\em Advances in neural information processing systems}, 30, 2017.

\bibitem{huang2021multimodal}
Xun Huang, Arun Mallya, Ting-Chun Wang, and Ming-Yu Liu.
\newblock Multimodal conditional image synthesis with product-of-experts gans.
\newblock {\em arXiv preprint arXiv:2112.05130}, 3, 2021.

\bibitem{huang2022multimodal}
Xun Huang, Arun Mallya, Ting-Chun Wang, and Ming-Yu Liu.
\newblock Multimodal conditional image synthesis with product-of-experts gans.
\newblock In {\em European Conference on Computer Vision}, pages 91--109.
  Springer, 2022.

\bibitem{isola2017image}
Phillip Isola, Jun-Yan Zhu, Tinghui Zhou, and Alexei~A Efros.
\newblock Image-to-image translation with conditional adversarial networks.
\newblock In {\em Proceedings of the IEEE conference on computer vision and
  pattern recognition}, pages 1125--1134, 2017.

\bibitem{jenkins1993importance}
DL Jenkins and RR Martin.
\newblock The importance of free-hand sketching in conceptual design: automatic
  sketch input.
\newblock In {\em International Design Engineering Technical Conferences and
  Computers and Information in Engineering Conference}, volume 11702, pages
  115--128. American Society of Mechanical Engineers, 1993.

\bibitem{kampelmuhlerP2020synthesizing}
Moritz Kampelm{\"{u}}hler and Axel Pinz.
\newblock Synthesizing human-like sketches from natural images using a
  conditional convolutional decoder.
\newblock In {\em {IEEE} Winter Conference on Applications of Computer Vision,
  {WACV} 2020, Snowmass Village, CO, USA, March 1-5, 2020}, pages 3192--3200.
  {IEEE}, 2020.

\bibitem{karras2017progressive}
Tero Karras, Timo Aila, Samuli Laine, and Jaakko Lehtinen.
\newblock Progressive growing of gans for improved quality, stability, and
  variation.
\newblock {\em arXiv preprint arXiv:1710.10196}, 2017.

\bibitem{karras2019style}
Tero Karras, Samuli Laine, and Timo Aila.
\newblock A style-based generator architecture for generative adversarial
  networks.
\newblock In {\em Proceedings of the IEEE/CVF conference on computer vision and
  pattern recognition}, pages 4401--4410, 2019.

\bibitem{rethinking2022media}
Hye-Kyung Lee.
\newblock Rethinking creativity: creative industries, ai and everyday
  creativity.
\newblock {\em Media, Culture \& Society}, pages Volume 44, Issue 3, 2022.

\bibitem{li2019photo}
Mengtian Li, Zhe~L. Lin, Radom{\'{\i}}r Mech, Ersin Yumer, and Deva Ramanan.
\newblock Photo-sketching: Inferring contour drawings from images.
\newblock In {\em {IEEE} Winter Conference on Applications of Computer Vision,
  {WACV} 2019, Waikoloa Village, HI, USA, January 7-11, 2019}, pages
  1403--1412. {IEEE}, 2019.

\bibitem{li2017free}
Yi Li, Yi{-}Zhe Song, Timothy~M. Hospedales, and Shaogang Gong.
\newblock Free-hand sketch synthesis with deformable stroke models.
\newblock {\em Int. J. Comput. Vis.}, 122(1):169--190, 2017.

\bibitem{lin2014microsoft}
Tsung-Yi Lin, Michael Maire, Serge Belongie, James Hays, Pietro Perona, Deva
  Ramanan, Piotr Doll{\'a}r, and C~Lawrence Zitnick.
\newblock Microsoft coco: Common objects in context.
\newblock In {\em European conference on computer vision}, pages 740--755.
  Springer, 2014.

\bibitem{liu2020sketch}
Bingchen Liu, Kunpeng Song, Yizhe Zhu, and Ahmed Elgammal.
\newblock Sketch-to-art: Synthesizing stylized art images from sketches.
\newblock In {\em Proceedings of the Asian Conference on Computer Vision},
  2020.

\bibitem{miller1995wordnet}
George~A Miller.
\newblock Wordnet: a lexical database for english.
\newblock {\em Communications of the ACM}, 38(11):39--41, 1995.

\bibitem{muhammad2018learning}
Umar~Riaz Muhammad, Yongxin Yang, Yi{-}Zhe Song, Tao Xiang, and Timothy~M.
  Hospedales.
\newblock Learning deep sketch abstraction.
\newblock In {\em 2018 {IEEE} Conference on Computer Vision and Pattern
  Recognition, {CVPR} 2018, Salt Lake City, UT, USA, June 18-22, 2018}, pages
  8014--8023. Computer Vision Foundation / {IEEE} Computer Society, 2018.

\bibitem{papineni2002bleu}
Kishore Papineni, Salim Roukos, Todd Ward, and Wei-Jing Zhu.
\newblock Bleu: a method for automatic evaluation of machine translation.
\newblock In {\em Proceedings of the 40th annual meeting of the Association for
  Computational Linguistics}, pages 311--318, 2002.

\bibitem{pinterest}
Inc. Pinterest.
\newblock Pinterest: Discover recipes, home ideas, style inspiration and other
  ideas to try., 2022.

\bibitem{ramesh2021zero}
Aditya Ramesh, Mikhail Pavlov, Gabriel Goh, Scott Gray, Chelsea Voss, Alec
  Radford, Mark Chen, and Ilya Sutskever.
\newblock Zero-shot text-to-image generation.
\newblock In {\em International Conference on Machine Learning}, pages
  8821--8831. PMLR, 2021.

\bibitem{rombach2022high}
Robin Rombach, Andreas Blattmann, Dominik Lorenz, Patrick Esser, and Bj{\"o}rn
  Ommer.
\newblock High-resolution image synthesis with latent diffusion models.
\newblock In {\em Proceedings of the IEEE/CVF Conference on Computer Vision and
  Pattern Recognition}, pages 10684--10695, 2022.

\bibitem{salimans2016improved}
Tim Salimans, Ian Goodfellow, Wojciech Zaremba, Vicki Cheung, Alec Radford, and
  Xi Chen.
\newblock Improved techniques for training gans.
\newblock {\em Advances in neural information processing systems}, 29, 2016.

\bibitem{sangkloy2016sketchy}
Patsorn Sangkloy, Nathan Burnell, Cusuh Ham, and James Hays.
\newblock The sketchy database: learning to retrieve badly drawn bunnies.
\newblock {\em ACM Transactions on Graphics (TOG)}, 35(4):1--12, 2016.

\bibitem{schaldenbrand2022styleclipdraw}
Peter Schaldenbrand, Zhixuan Liu, and Jean Oh.
\newblock Styleclipdraw: Coupling content and style in text-to-drawing
  translation.
\newblock {\em arXiv preprint arXiv:2202.12362}, 2022.

\bibitem{schuhmann2022laion}
Christoph Schuhmann, Romain Beaumont, Richard Vencu, Cade Gordon, Ross
  Wightman, Mehdi Cherti, Theo Coombes, Aarush Katta, Clayton Mullis, Mitchell
  Wortsman, et~al.
\newblock Laion-5b: An open large-scale dataset for training next generation
  image-text models.
\newblock {\em arXiv preprint arXiv:2210.08402}, 2022.

\bibitem{schuhmann2021laion}
Christoph Schuhmann, Richard Vencu, Romain Beaumont, Robert Kaczmarczyk,
  Clayton Mullis, Aarush Katta, Theo Coombes, Jenia Jitsev, and Aran
  Komatsuzaki.
\newblock Laion-400m: Open dataset of clip-filtered 400 million image-text
  pairs.
\newblock {\em arXiv preprint arXiv:2111.02114}, 2021.

\bibitem{sharma2018conceptual}
Piyush Sharma, Nan Ding, Sebastian Goodman, and Radu Soricut.
\newblock Conceptual captions: A cleaned, hypernymed, image alt-text dataset
  for automatic image captioning.
\newblock In {\em Proceedings of the 56th Annual Meeting of the Association for
  Computational Linguistics (Volume 1: Long Papers)}, pages 2556--2565, 2018.

\bibitem{szegedy2016rethinking}
Christian Szegedy, Vincent Vanhoucke, Sergey Ioffe, Jon Shlens, and Zbigniew
  Wojna.
\newblock Rethinking the inception architecture for computer vision.
\newblock In {\em Proceedings of the IEEE conference on computer vision and
  pattern recognition}, pages 2818--2826, 2016.

\bibitem{vinker2022clipasso}
Yael Vinker, Ehsan Pajouheshgar, Jessica~Y Bo, Roman~Christian Bachmann,
  Amit~Haim Bermano, Daniel Cohen-Or, Amir Zamir, and Ariel Shamir.
\newblock Clipasso: Semantically-aware object sketching.
\newblock {\em arXiv preprint arXiv:2202.05822}, 2022.

\bibitem{vojtovivc2015creative}
S Vojtovi{\v{c}}.
\newblock Creative industry as a sector of the new economy.
\newblock {\em Actual problems of modern economy development}, page~22, 2015.

\bibitem{wang2018high}
Ting-Chun Wang, Ming-Yu Liu, Jun-Yan Zhu, Andrew Tao, Jan Kautz, and Bryan
  Catanzaro.
\newblock High-resolution image synthesis and semantic manipulation with
  conditional gans.
\newblock In {\em Proceedings of the IEEE conference on computer vision and
  pattern recognition}, pages 8798--8807, 2018.

\bibitem{zhang2021m6}
Zhu Zhang, Jianxin Ma, Chang Zhou, Rui Men, Zhikang Li, Ming Ding, Jie Tang,
  Jingren Zhou, and Hongxia Yang.
\newblock M6-ufc: Unifying multi-modal controls for conditional image
  synthesis.
\newblock {\em arXiv preprint arXiv:2105.14211}, 2021.

\bibitem{zhu2017unpaired}
Jun-Yan Zhu, Taesung Park, Phillip Isola, and Alexei~A Efros.
\newblock Unpaired image-to-image translation using cycle-consistent
  adversarial networks.
\newblock In {\em Proceedings of the IEEE international conference on computer
  vision}, pages 2223--2232, 2017.

\bibitem{zou2018sketchyscene}
Changqing Zou, Qian Yu, Ruofei Du, Haoran Mo, Yi-Zhe Song, Tao Xiang, Chengying
  Gao, Baoquan Chen, and Hao Zhang.
\newblock Sketchyscene: Richly-annotated scene sketches.
\newblock In {\em Proceedings of the european conference on computer vision
  (ECCV)}, pages 421--436, 2018.

\end{thebibliography}
}

\end{document}